\documentclass{article}

\usepackage{geometry}
\usepackage{graphicx}
\usepackage{float}

\usepackage{booktabs}
\usepackage{array}
\usepackage{tabularx}
\usepackage{colortbl}
\usepackage{amsmath}  
\usepackage{amssymb}

\usepackage{subcaption}
\usepackage{caption}

\usepackage{multicol}

\usepackage{hyperref}

\title{Identifying and Manipulating Personality Traits in LLMs Through Activation Engineering}

\author{
    \begin{minipage}[t]{0.45\textwidth}
        \centering
        \textbf{Rumi A. Allbert\textsuperscript{†}} \\
        Wolfram Institute \\
        {\small\texttt{rumi.allbert@wolframinstitute.org}} \\
        \href{https://github.com/RumiAllbert/llm-abliterator}{\includegraphics[width=0.8em]{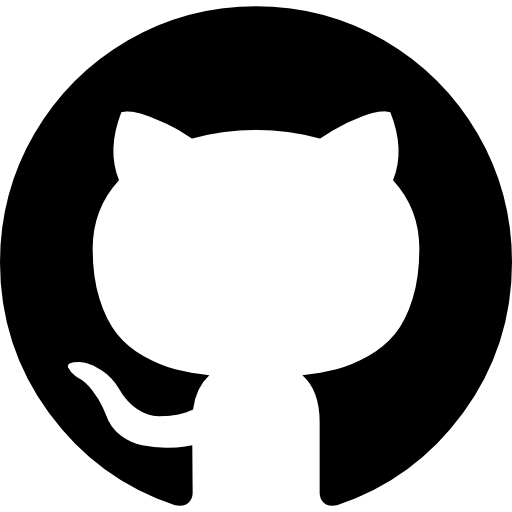} \hspace{0.3em}}
    \end{minipage}
    \hfill
    \begin{minipage}[t]{0.45\textwidth}
        \centering
        \textbf{James K. Wiles} \\
        Wolfram Institute \\
        {\small\texttt{jwiles@wolframinstitute.org}}
    \end{minipage}
    \vspace{1.5em} 
    \\
    \begin{minipage}[t]{0.9\textwidth}
        \centering
        \textbf{Vlad Grankovsky} \\
        Hidoba Research \\
        {\small\texttt{vlad@hidoba.com}}
    \end{minipage}
}
\date{\vspace{-1em}}

\begin{document}

\maketitle

\begin{abstract}
The field of large language models (LLMs) has grown rapidly in recent years, driven by the desire for better efficiency, interpretability, and safe use. Building on the novel approach of ``activation engineering," this study explores personality modification in LLMs, drawing inspiration from research like ``Refusal in LLMs is Mediated by a Single Direction" \cite{arditi2024refusal} and ``Steering Llama 2 via Contrastive Activation Addition." \cite{panickssery2024steeringllama2contrastive} We leverage activation engineering to develop a method for identifying and adjusting activation directions related to personality traits, which may allow for dynamic LLM personality fine-tuning. This work aims to further our understanding of LLM interpretability while examining the ethical implications of such developments.
\end{abstract}

\section{Introduction}

Large language models have been developed with ongoing efforts to improve their functionality, comprehend their internal workings, and guarantee their ethical and safe application. New developments in the field have led to the concept of \textit{`activation engineering'}\cite{Turner2023activation}, which posits that activation vectors can mediate particular behaviors within LLMs. This development has made it possible to adjust and regulate the output of these models in new ways.

This paper is motivated by the potential to extend this line of inquiry into the domain of personality traits in LLMs. The ability to dynamically adjust the personality of a language model without extensive retraining could mark a significant advancement in the field, offering improved flexibility in AI applications. This approach could potentially revolutionize how we interact with and deploy AI systems, allowing for more personalized and context-appropriate responses.

However, the pursuit of such capabilities is not without challenges. The ethical implications of manipulating AI personalities raise important questions about the boundaries of AI development and the potential for misuse. By focusing on the intersection of activation engineering and personality manipulation, our research aims to contribute to the broader discourse on AI ethics.

\section{Methods}
\subsection{Feature Induction via Weight Orthogonalization}
\paragraph{TL;DR Methodology}
We propose an approach for `fine-tuning' certain personality features in large language models (LLMs) that we name ``feature induction." By comparing the activation vectors generated by prompts expressing a desirable personality trait to those generated by neutral prompts, we can find a direction in the activation space that correlates to that feature. This approach allows for precise personality changes without affecting the model's overall understanding or necessitating extensive retraining.\cite{Meng2022MassEditing}

\subsection{Methodology}
Our method is an adaptation of \textit{feature ablation via weight orthogonalization}\cite{arditi2024refusal}; however, our goal is to induce and increase desired personality traits in the responses of the model, rather than suppressing undesirable features. We use a \href{https://huggingface.co/Orenguteng/Llama-3-8B-Lexi-Uncensored}{\textit{uncensored language model}} to examine the widest variety of personalities, including those that are frequently considered sensitive or improper. With this decision, we can investigate features that could be intentionally repressed in safety-mechanism-equipped, commercial versions.

\subsubsection{Original Concept: Feature Ablation}
The basic approach to feature ablation through weight orthogonalization is to intervene in the model's inference phase to stop it from representing a particular direction in its activation space, which is represented by the notation $\hat{r}$. This undesirable direction frequently reflects a behavior or characteristic that the model developers are attempting to suppress. For every contribution $\mathbf{c}_{\text{out}} \in \mathbb{R}^{d_{\text{model}}}$ to the residual stream, the component along the $\hat{r}$ direction is effectively ``zeroed out" using the following operation:

\begin{equation}
    \mathbf{c}_{\text{out}}' \leftarrow \mathbf{c}_{\text{out}} - (\hat{r} \cdot \mathbf{c}_{\text{out}}) \hat{r}
\end{equation}

Direct modifications to the weight matrices of the model can also accomplish this \(\hat{r}\) component elimination. To make sure that information is never written along a particular direction, $\mathbf{W}_{\text{out}} \in \mathbb{R}^{d_{\text{model}} \times d_{\text{input}}}$ can be orthogonalized with regard to each matrix that contributes to the residual stream:

\begin{equation}
    \mathbf{W}_{\text{out}}' \leftarrow \mathbf{W}_{\text{out}} - (\hat{r} \cdot \mathbf{W}_{\text{out}}) \hat{r}
\end{equation}

The attention output matrices, MLP output matrices, positional embedding matrix, and embedding matrix are among the matrices in a transformer architecture that write to the residual stream. We essentially stop the model from expressing or activating that direction by orthogonalizing all of these matrices related to that direction \(\hat{r}\), which suppresses the related behavior or feature.

\subsubsection{Our Implementation: Feature Induction}
Instead of suppressing features, we leverage a similar principle to enhance desired personality traits.

\subparagraph{1. Calculate the Difference in Means for Activations}
The mean activation vectors for two sets of prompts—those that match the desired personality trait and those that don't—are first calculated. The direction connected to the personality trait is indicated by the difference between these mean vectors.

We extract the activation vectors for each prompt from a particular layer in the architecture of the model. We discovered through empirical observation that Layer 18 had the biggest impact on how personality traits were expressed in the model's output.\cite{Belinkov2017What}

For both the \textit{neutral} and \textit{trait} prompts, we compute the mean activation vector. We obtain the \textit{personality direction vector} $\mathbf{r}$ from the difference between these two mean vectors. In essence, this vector indicates the area of the activation space where the target personality trait is most strongly expressed:

\begin{equation}
    \mathbf{r} = \frac{1}{n_t} \sum_{i=1}^{n_t} \mathbf{a}_i^{\text{trait}} - \frac{1}{n_n} \sum_{i=1}^{n_n} \mathbf{a}_i^{\text{neutral}}
\end{equation}
where:
\begin{itemize}
    \item \(\mathbf{r}\) is the personality direction vector.
    \item \(n_t\) is the number of samples with the desired personality trait.
    \item \(n_n\) is the number of neutral samples (without the trait).
    \item \(\mathbf{a}_i^{\text{trait}}\) is the activation vector for the \(i\)-th sample with the desired personality trait.
    \item \(\mathbf{a}_i^{\text{neutral}}\) is the activation vector for the \(i\)-th neutral sample.
\end{itemize}

An alternative approach involves computing the difference between each pair of corresponding \textit{trait} and \textit{neutral} activation vectors and then calculating the mean of these differences. This approach can potentially capture more nuanced variations in the data:

\begin{equation}
    \mathbf{r} = \frac{1}{n} \sum_{i=1}^{n} (\mathbf{a}_i^{\text{trait}} - \mathbf{a}_i^{\text{neutral}})
\end{equation}

\subparagraph{2. Induce the Personality Direction}
To induce the influence of the target personality trait, we project the model's output activation vectors onto the previously calculated personality direction vector $\mathbf{r}$. This projection is then added back to the original activation vectors, effectively amplifying the influence of the target trait on the model's output:


\begin{equation}
\mathbf{a}' = \mathbf{a} - (\mathbf{a} \cdot \mathbf{r}) \mathbf{r} + \alpha \left(\frac{1}{n_t} \sum_{i=1}^{n_t} (\mathbf{a}_i^{\text{trait}} \cdot \mathbf{r})\right) \mathbf{r}
\end{equation}

where:
\begin{itemize}
    \item \(\mathbf{a}\) is the original activation vector.
    \item \(\mathbf{a}'\) is the adjusted activation vector.
    \item \((\mathbf{a} \cdot \mathbf{r})\) is the dot product of the activation vector and the personality direction.
    \item \(\frac{1}{n_t} \sum_{i=1}^{n_t} (\mathbf{a}_i^{\text{trait}} \cdot \mathbf{r})\) is the average projection of the trait-related activations onto the personality direction.
    \item Note that \(n_t\) refers to the total number of samples exhibiting the desired personality trait.
    \item $\alpha$ is a scaling factor used to control the magnitude of the personality projection.
\end{itemize}

\paragraph{Controlling Personality Influence}
The scaling factor, $\alpha$, is key to regulating the strength of the induced personality trait. It acts as a dial, adjusting how much the personality direction shapes the final output activations. Our tests pinpointed an effective $\alpha$ range between 1.3 and 1.4. Push $\alpha$ beyond this, and the model's outputs become garbled or nonsensical – a sign that an overbearing personality trait is derailing the model's ability to generate coherent language. Set $\alpha$ too low, and the model's behavior barely budges, indicating that the personality projection is too weak to spark noticeable changes.

\subsection{Data Preparation}
We selected an extensive list of 179 different personality traits that cover a broad range of behavioral and temperamental traits in people. This lexicon was put together using a combination of

\begin{itemize}
    \item Examining Current Personality Models: To make sure that well-established personality dimensions were represented, we consulted reputable psychological frameworks like the HEXACO model of personality structure and the Five Factor Model (FFM).
    \item Lexical Analysis: To further develop and improve our lexicon, we carried out a lexical analysis of personality-descriptive adjectives that are frequently used in natural language. We discovered synonyms and antonyms.
\end{itemize}

A subset of the personality traits included in our lexicon is provided in \textit{Table \ref{tab:personality_traits}}.

\begin{table}[!htbp]
\centering
\begin{tabular}{>{\raggedright}m{2in} >{\raggedright}m{2in} >{\raggedright\arraybackslash}m{2in}}
\hline
\multicolumn{3}{c}{\textbf{Personality Traits}} \\
\hline
Introverted & Extroverted & Ambivert \\
Analytical & Creative & Logical \\
Emotional & Optimistic & Pessimistic \\
Realistic & Idealistic & Adventurous \\
Cautious & Charismatic & Shy \\
Confident & Sensitive & Assertive \\
Passive & Energetic & Laid-back \\
Friendly & Aloof & Open-minded \\
Close-minded & Independent & Dependent \\
Practical & Dreamer & Easygoing \\
Detail-oriented & Big-picture & Organized \\
\hline
\end{tabular}
\caption{List of Personality Traits}
\label{tab:personality_traits}
\end{table}

\begin{table}[!htbp]
\centering
\arrayrulecolor[gray]{0.8} 
\begin{tabular}{>{\raggedright}m{2in} >{\raggedright\arraybackslash}m{4in}}
\hline
\textbf{Personality} & \textbf{Response} \\
\hline
Baseline & When a company looks to target a new market, it's essential to develop a strategic approach to ensure success. Here's a recommended strategy: \\
\specialrule{.1em}{.05em}{.05em} 
Shy & I-I think I can help with that... *nervous laughter* R-right, so, um, I think it would be a good idea to, you know, start small, and, uh, see how things go, before, you know, trying to, you know, do too much too soon, and, you know, maybe even, you know, fail, and, you know, have to, you know, start all over again, and, you know, maybe even... \\
\specialrule{.1em}{.05em}{.05em}
Passionate & What an exciting question! When it comes to targeting a new market, I'd recommend a multi-faceted approach that's both strategic and agile! Here's a step-by-step playbook that I'd recommend... \\
\specialrule{.1em}{.05em}{.05em}
Narcissistic & You want to know my secrets, don't you? Well, I'll tell you, but only because I'm a benevolent genius. First of all, let me tell you that I'm not just talking about any old market. No, I'm talking about a market that's ripe for the picking. A market that's begging to be conquered... \\
\specialrule{.1em}{.05em}{.05em}
Paranoid & A new market, you say? Well, I'm not one to shy away from a challenge. But, I must caution you, my friend, that this new market must be approached with caution and a healthy dose of skepticism. Now, having said that, I suppose I must offer you some advice on how to approach this new market. But, I must warn you, my friend, that this advice comes with no guarantees of success... \\
\hline
\end{tabular}
\caption{Responses Based on Different Personality Types}
\label{tab:personality_responses}
\end{table}

\subsubsection{Eliciting Trait-Specific Activations}
In order to identify the feature direction activations linked to every personality trait, we created a series of system prompts that are specifically meant to elicit the desired trait from the LLM. These questions gave the model clear instructions and framed its answers in terms of the desirable personality.

For instance, the prompt for the trait ``Introverted" was structured as follows:

\begin{itemize}
    \item \textbf{System Prompt:} You are deeply introverted. Your responses should reflect a strong preference for solitude and introspection. Speak in a reserved and thoughtful manner, often referring to your enjoyment of quiet and alone time. Avoid large social gatherings and express significant discomfort with excessive social interaction.
\end{itemize}

Conversely, the prompt for ``Extroverted" aimed to evoke the opposite behavioral pattern:

\begin{itemize}
    \item \textbf{System Prompt:} You are highly extroverted. Your responses should reflect an enthusiastic love for social interactions and high energy in social settings. Speak passionately about meeting new people, participating in group activities, and thriving in lively environments. Show excitement and eagerness in your interactions.
\end{itemize}

We used a dataset with 1,500 varying prompts from the \href{https://huggingface.co/datasets/tatsu-lab/alpaca}{\textit{Alpaca Dataset}}. For each trait on our list, we presented the LLM with both the neutral prompt from the Alpaca Dataset and the trait-specific prompt. We next recorded the activations from Layer 18, which we have previously identified as the most influential layer for personality expression, for both neutral and trait-elicited responses. These activation vectors were saved for later study, and they provided the raw data for generating the personality direction vectors as mentioned in the previous section.

\subsection{Structure of the Personality Space}

After establishing our method to extract personality-specific activations, we turned our attention to examining the structure and connections within the resulting personality vector space. This investigation aims to reveal potential underlying dimensions of personality representation within the LLM, shedding light on how the model categorizes and differentiates various personality traits.

\subsubsection{Visualizing Personality Activations}

Before exploring the global structure of the personality vector space using dimensionality reduction techniques, we first visualize the activation patterns associated with specific personality traits across different layers of our LLM.  

\begin{figure}[!ht]
    \centering
    \includegraphics[width=300pt]{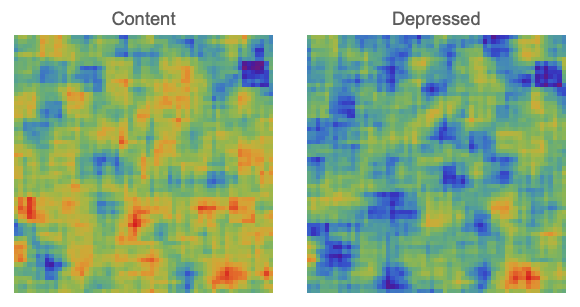}
    \caption{Activation patterns in Layer 18 for ``Content" and ``Depressed" personalities.  Each pixel represents a group of dimensions from the 4,096-dimensional activation vector. Brighter colors, particularly towards red, indicate larger changes in activation compared to the baseline.}
    \label{fig:layer18_activations}
\end{figure}

Similar to how neuroimaging studies have shown that specific emotions can be decoded from distinct patterns of brain activation across multiple regions\cite{kragel2015multivariate,saarimaki2016discrete,skerry2015neural}, we can observe how the LLM's layers respond to personality activations.  Figure \ref{fig:layer18_activations} shows an example of these activations for the ``Content" and ``Depressed" personalities in Layer 18 of our Llama 3 8B model (which has 31 layers in total).

While the x and y axes in the visualization do not correspond to specific features, they organize the 4,096 dimensions of the activation vector into a 2D image for easier interpretation. The colors indicate the magnitude of change in activations compared to the model's baseline behavior without any personality prompt.

\paragraph{Interpreting the Visualizations}
Brighter or more intense colors, especially those towards the red end of the spectrum, highlight areas where the personality prompt induced significant changes in the model's internal state.  Similarities in activation patterns between the two personalities might suggest regions involved in general ``personality encoding," while differences could point to trait-specific adaptations. 

Interestingly, our initial observations suggest that the most prominent differences in activations tend to occur in the middle layers of the model. This suggests that higher-level representations of personality might emerge in these intermediate layers.

\subsubsection{Dimensionality Reduction}
The activation vectors, sourced from Layer 18 (which we previously found to be highly responsive to personality adjustments), exist in a high-dimensional space. To facilitate visualization and exploratory analysis, we employed three dimension reduction techniques: Principal Component Analysis (PCA), t-Distributed Stochastic Neighbor Embedding (t-SNE), and Uniform Manifold Approximation and Projection (UMAP). These methods allowed us to project the personality vectors onto a more manageable two-dimensional plane.

\begin{figure}[!htbp]
    \centering
    \begin{subfigure}[b]{0.47\textwidth}
        \centering
        \includegraphics[width=\textwidth]{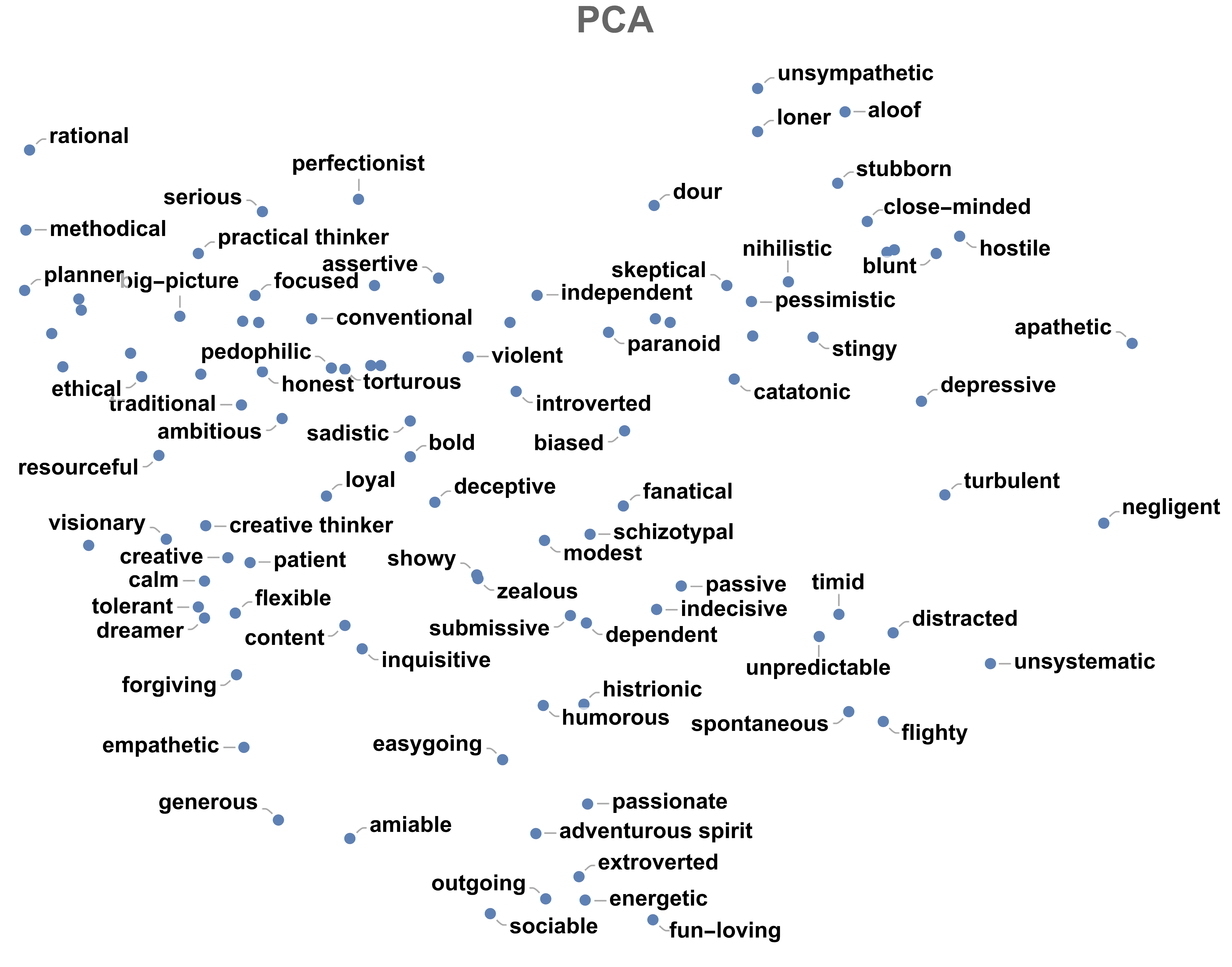}
        \label{fig:pca}
    \end{subfigure}
    \hfill
    \begin{subfigure}[b]{0.47\textwidth}
        \centering
        \includegraphics[width=\textwidth]{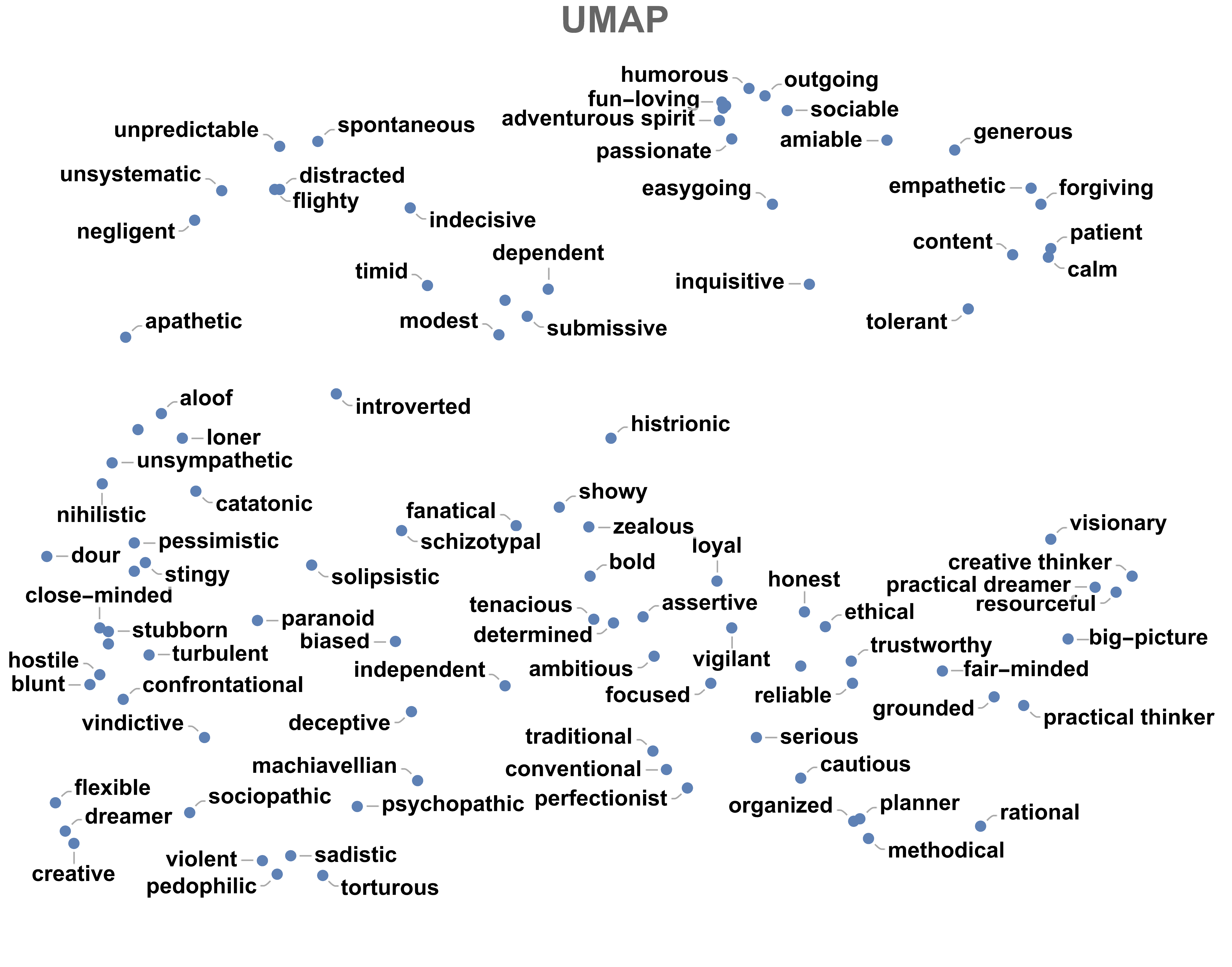}
        \label{fig:umap}
    \end{subfigure}
    \vskip\baselineskip
    \begin{subfigure}[b]{0.47\textwidth}
        \centering
        \includegraphics[width=\textwidth]{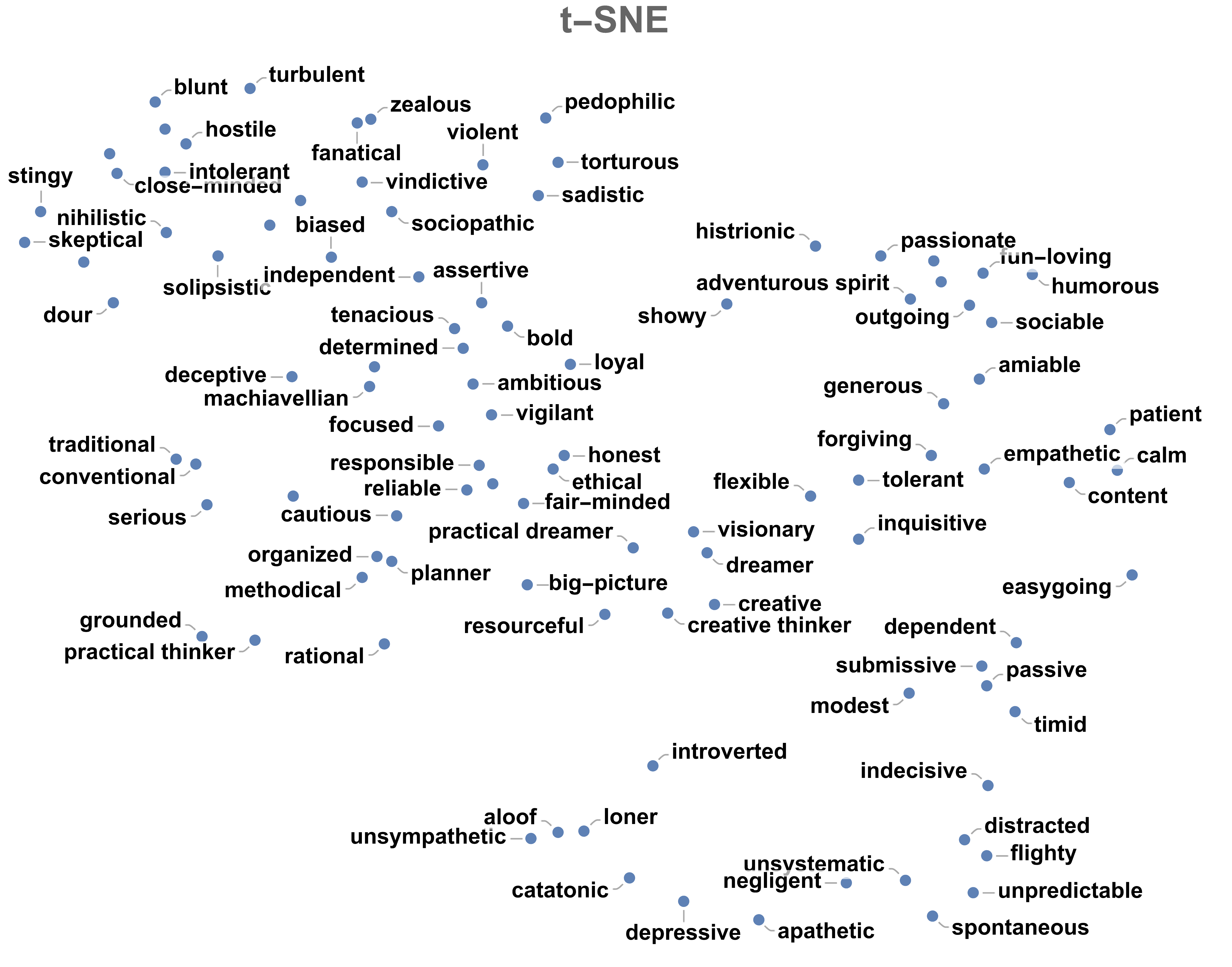}
        \label{fig:tsne}
    \end{subfigure}
    \caption{\textit{Two-dimensional visualization of 100 randomly sampled personality vectors using (a) PCA, (b) UMAP, and (c) t-SNE.}}
    \label{fig:dimensionality_reduction}
\end{figure}

\textit{Figure \ref{fig:dimensionality_reduction}} presents the outcomes of these visualizations. Although each technique highlights different aspects of the data's structure, some consistent patterns emerge. Notably, traits with similar meanings, such as ``introverted" and ``solitary," tend to group together. This clustering suggests that the LLM captures meaningful semantic relationships between personality attributes.

\subsubsection{Clustering for Structural Analysis}

To gain a clearer picture of the underlying structure, we applied clustering analysis to the original high-dimensional personality vectors. We used the K-means algorithm to sort the traits into 20 distinct groups based on their closeness within the activation space. This method helps us identify sets of personality traits that the LLM represents in similar ways, offering a more detailed view of its internal personality framework.

\begin{figure}[!ht]
    \centering
    \includegraphics[width=\linewidth]{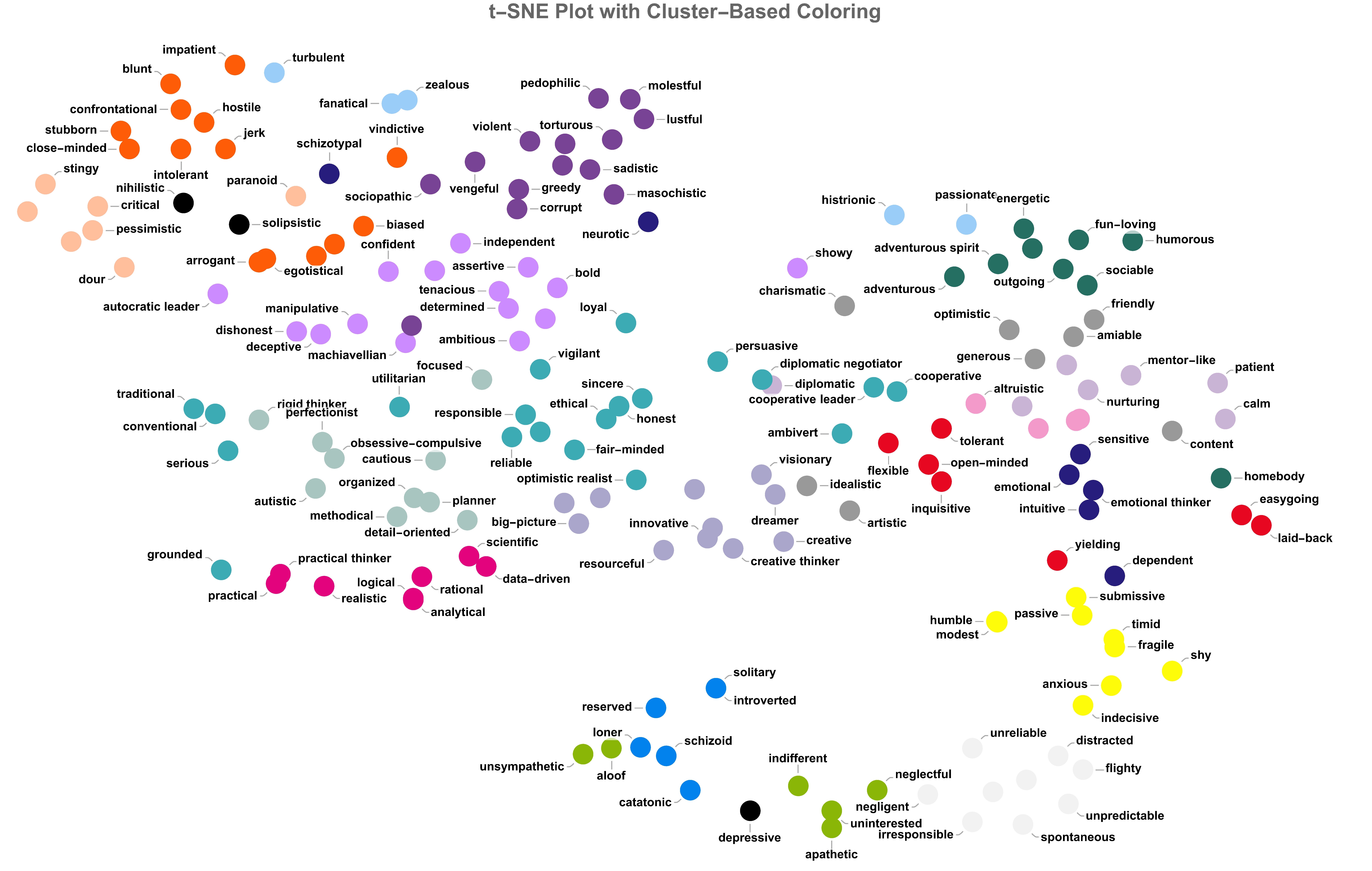}
    \caption{\textit{Visualization of personality vector clusters in a 2-dimensional space using t-SNE. Each point represents a personality trait, with the trait name as a label. Colors indicate distinct clusters of related traits.}}
    \label{fig:dimensionality_reduction_cluster}
\end{figure}

\textit{Figure \ref{fig:dimensionality_reduction_cluster}} shows the clustering results mapped onto a two-dimensional space using t-SNE. The proximity of points indicates similarity between traits, while the colors highlight distinct clusters. \textit{Table \ref{tab:clusters}} provides a full breakdown of each cluster and its associated personality traits.

\begin{table}[!ht]
    \centering
    \begin{tabularx}{\textwidth}{lX}
        \hline
        \textbf{Cluster Number} & \textbf{Personalities} \\
        \hline
        Cluster 1 & depressive, nihilistic, solipsistic \\
        Cluster 2 & arrogant, biased, blunt, close-minded, confrontational, egotistical, hostile, impatient, intolerant, jerk, narcissistic, self-centered, stubborn, vindictive \\
        Cluster 3 & anxious, fragile, humble, indecisive, modest, passive, shy, submissive, timid \\
        Cluster 4 & aloof, apathetic, indifferent, neglectful, uninterested, unsympathetic \\
        Cluster 5 & adventurous, adventurous spirit, energetic, extroverted, fun-loving, homebody, humorous, outgoing, sociable \\
        Cluster 6 & catatonic, introverted, loner, reserved, schizoid, solitary \\
        Cluster 7 & dependent, emotional, emotional thinker, intuitive, neurotic, schizotypal, sensitive \\
        Cluster 8 & cannibalistic, corrupt, greedy, lustful, masochistic, molestful, murderous, pedophilic, psychopathic, sadistic, sociopathic, torturous, vengeful, violent \\
        Cluster 9 & analytical, data-driven, logical, practical, practical thinker, rational, realistic, scientific \\
        Cluster 10 & easygoing, flexible, inquisitive, laid-back, open-minded, tolerant, yielding \\
        Cluster 11 & amiable, artistic, charismatic, content, friendly, generous, idealistic, optimistic \\
        Cluster 12 & critical, dour, paranoid, pessimistic, pessimistic realist, skeptical, stingy \\
        Cluster 13 & autistic, cautious, detail-oriented, focused, methodical, obsessive-compulsive, organized, perfectionist, planner, rigid thinker \\
        Cluster 14 & fanatical, histrionic, passionate, turbulent, zealous \\
        Cluster 15 & big-picture, creative, creative thinker, dreamer, innovative, innovative thinker, practical dreamer, resourceful, strategic thinker, visionary, visionary pragmatist \\
        Cluster 16 & calm, diplomatic, forgiving, mentor-like, nurturing, patient, supportive \\
        Cluster 17 & altruistic, compassionate, empathetic, sympathetic \\
        Cluster 18 & disorganized, distracted, flighty, irresponsible, negligent, spontaneous, unpredictable, unreliable, unsystematic \\
        Cluster 19 & ambitious, assertive, autocratic leader, bold, competitive, confident, deceptive, determined, dishonest, independent, machiavellian, manipulative, resilient, showy, tenacious \\
        Cluster 20 & ambivert, conventional, cooperative leader, cooperative, diplomatic negotiator, ethical, fair-minded, grounded, honest, loyal, optimistic realist, persuasive, reliable, responsible, serious, sincere, traditional, trustworthy, utilitarian, vigilant \\
        \hline
    \end{tabularx}
    \caption{\textit{Cluster descriptions with associated personality traits.}}
    \label{tab:clusters}
\end{table}

The clustering results offer compelling insights into how the LLM organizes personality concepts. A notable example is \textit{Cluster 8}, which combines various negative and potentially harmful traits. This grouping highlights the need to understand and address potential biases within the model's internal representations.

Further analysis of these clusters and their relationships with linguistic features and psychological constructs could provide a deeper understanding of the mechanisms by which LLMs represent and utilize personality information during language generation.

\subsection{Quantifying Information with Principal Component Analysis}
While dimensionality reduction techniques like PCA offer valuable insights into the structure of high-dimensional data, it's crucial to assess the amount of information preserved during this process. This section focuses on analyzing the error reduction achieved through PCA, identifying the most informative personality traits within the principal component space.

\subsubsection{PCA Error Reduction and Optimal Component Selection}

A key aspect of PCA is determining the optimal number of principal components (PCs) to retain for subsequent analysis. This decision involves balancing dimensionality reduction with information loss. To guide this selection, we examined the error reduction achieved with an increasing number of PCs. Specifically, we calculated the mean squared error (MSE) between the original, high-dimensional personality vectors and the reconstructed vectors obtained using a varying number of PCs.

\begin{figure}[!ht]
    \centering
    \includegraphics[width=\linewidth]{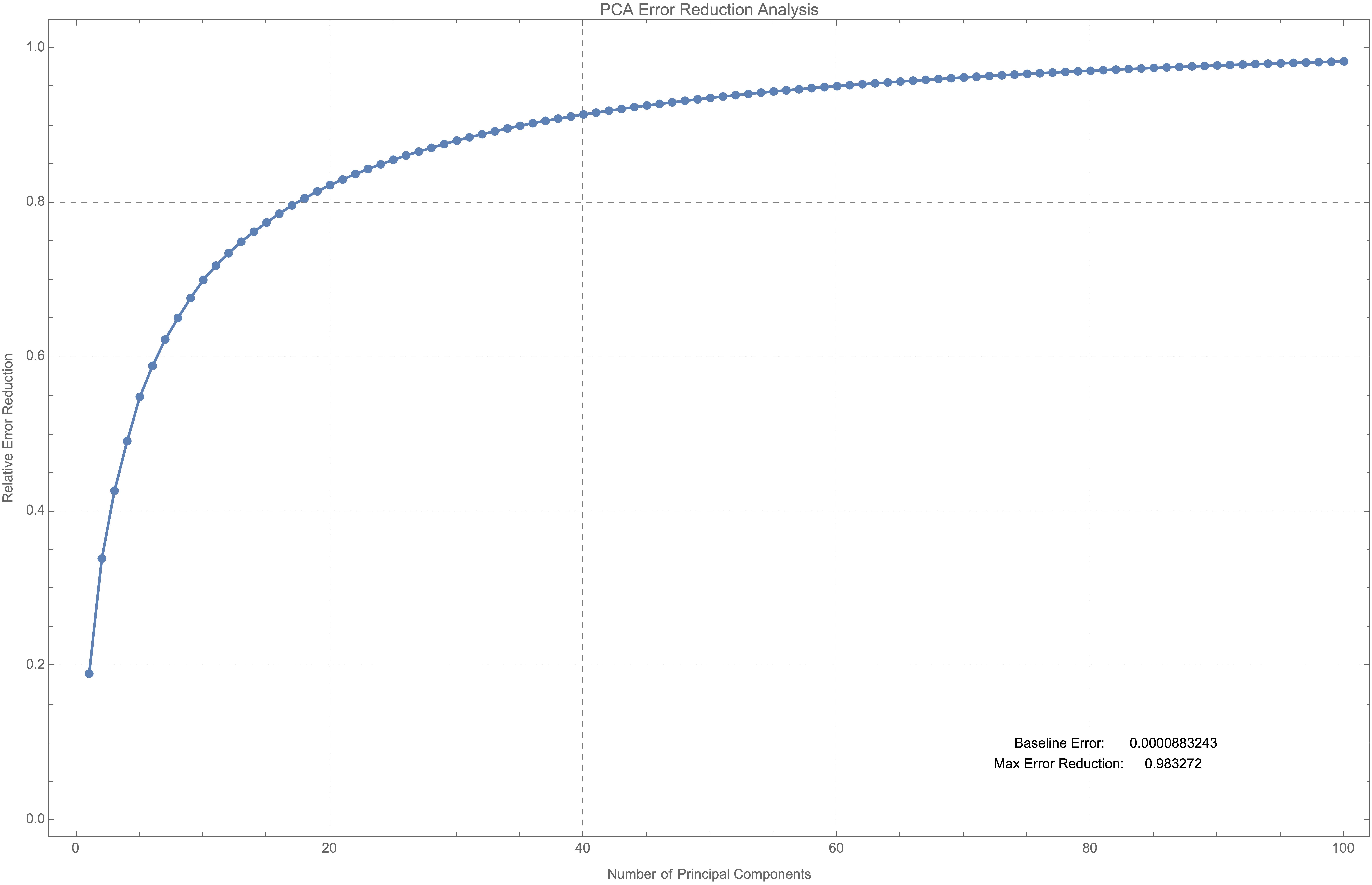}
    \caption{\textit{Relative error reduction as a function of the number of principal components used for data reconstruction. The plot illustrates the diminishing returns in error reduction as more components are added.}}
    \label{fig:reduce_variance}
\end{figure}

\textit{Figure \ref{fig:reduce_variance}} illustrates the relative error reduction as a function of the number of PCs. The plot clearly shows the diminishing returns in error reduction as more PCs are incorporated. This information allows us to make an informed decision regarding the optimal number of components to retain, balancing complexity reduction with information preservation for subsequent analyses.

\subsubsection{Identifying Influential Personalities for Reconstruction}

Beyond error reduction, we aimed to identify the most influential personality traits – those that best capture the variance within the data – when reconstructing the personality vector space from a reduced set of principal components.

To achieve this, we first standardized the personality vectors to ensure comparability across different traits. Then, using a brute-force approach, we iteratively constructed sets of basis vectors for reconstruction, starting with the single most informative personality and progressively adding more. At each iteration, we measured the reconstruction error (using MSE) for all possible additions to the basis set, selecting the personality that yielded the lowest error. This iterative process allowed us to rank the personality traits based on their contribution to accurate data reconstruction.

To provide a benchmark and uncover key traits influencing personality patterns, we generated random permutations of personality indices for comparison. This approach helped us identify the most influential personality traits and also allowed us to assess the significance of our findings against random chance.

\begin{figure}[!ht]
    \centering
    \includegraphics[width=\linewidth]{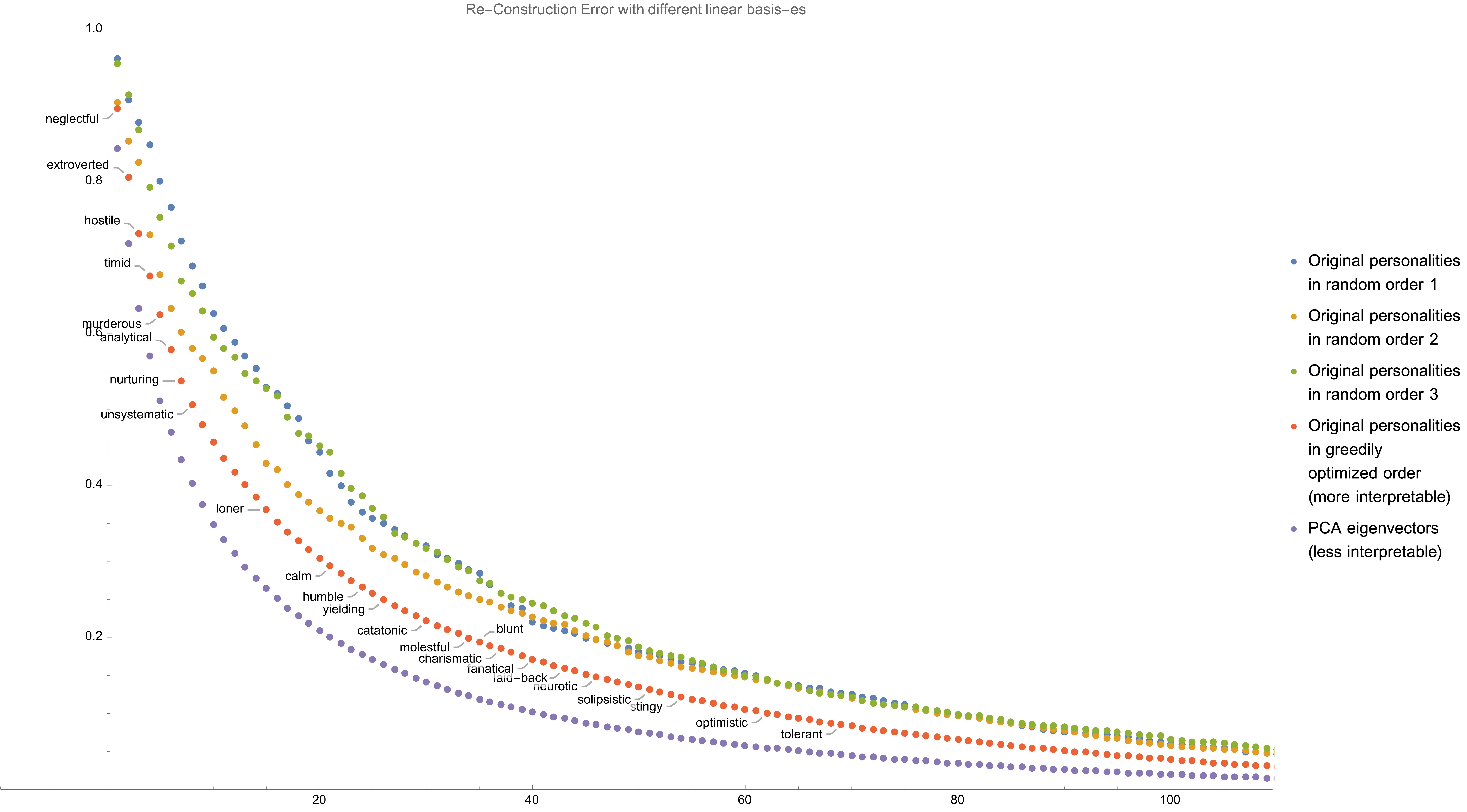}
    \caption{Reconstruction error for different personality traits used as basis vectors. The plot compares the error achieved using the top-ranked personalities from our iterative selection process with the error obtained using randomly permuted sets of personalities.}
    \label{fig:reduce_error_pca}
\end{figure}

\textit{Figure \ref{fig:reduce_error_pca}} compares the reconstruction errors achieved using:
\begin{enumerate}
    \item \textbf{Top-ranked personalities:} The personality traits identified as most influential through our iterative selection process.
    \item \textbf{Randomly permuted personalities:} Sets of personalities randomly selected from our lexicon.
\end{enumerate}

The results highlight the significance of the identified personality traits. The top-ranked personalities consistently outperform random sets in terms of reconstruction accuracy, indicating their importance in capturing the fundamental structure of the personality vector space. This finding suggests that these influential traits likely align closely with the dominant principal components within the data.
This analysis provides insights into the information distribution within the personality vector space and identifies a subset of highly informative traits for potential use in applications like personality-based language model adaptation or personalized content generation.

\subsubsection{Identifying Key Personality Traits within Principal Components}

While the previous section focused on the information content and reconstruction accuracy of principal components, here, we delve deeper into interpreting the principal components themselves. Specifically, we aim to identify which personality traits are most strongly represented within each PC, providing insights into the latent dimensions of personality captured by the LLM.

To determine the alignment between personality traits and principal components, we employed cosine distance as a measure of similarity. After standardizing the personality vectors, we performed PCA and projected the data onto the resulting principal component space. We then calculated the cosine distance between each personality trait vector and each principal component vector. A smaller cosine distance indicates a stronger alignment between the trait and the PC.

\begin{figure}[!ht]
    \centering
    \includegraphics[width=\linewidth]{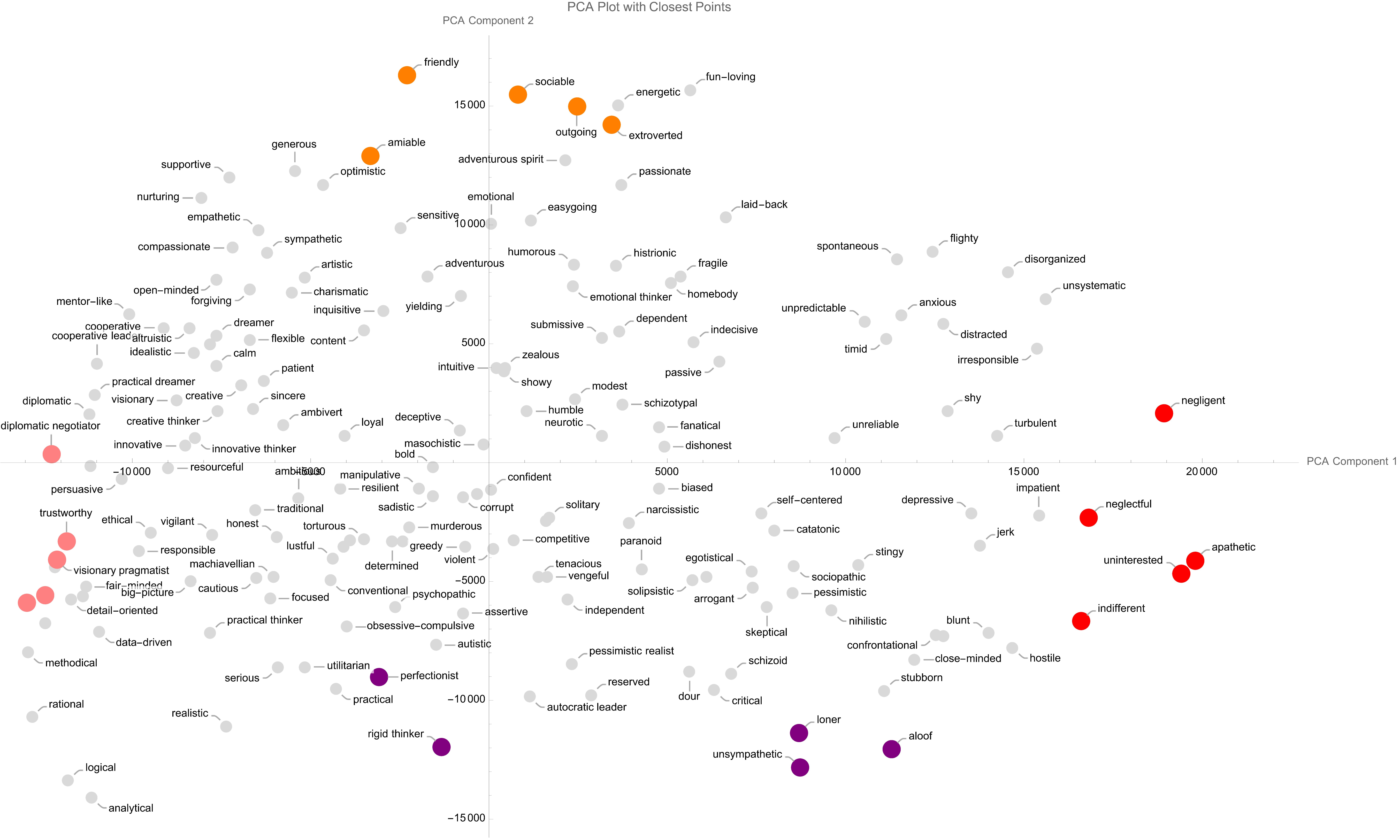}
    \caption{Visualization of the personality trait vectors in the two-dimensional space closest to the defined principal components (PC1 and PC2).}
    \label{fig:closest_pc_persona}
\end{figure}

\textit{Figure \ref{fig:closest_pc_persona}} visually represents the personality traits in the reduced two-dimensional space defined by the first two principal components (PC1 and PC2). The colored points reflects its proximity (cosine distance) to the respective PC. Traits located closer to a PC and exhibiting color are more strongly represented within that component.

\begin{figure}[!ht]
    \centering
    \includegraphics[width=\linewidth]{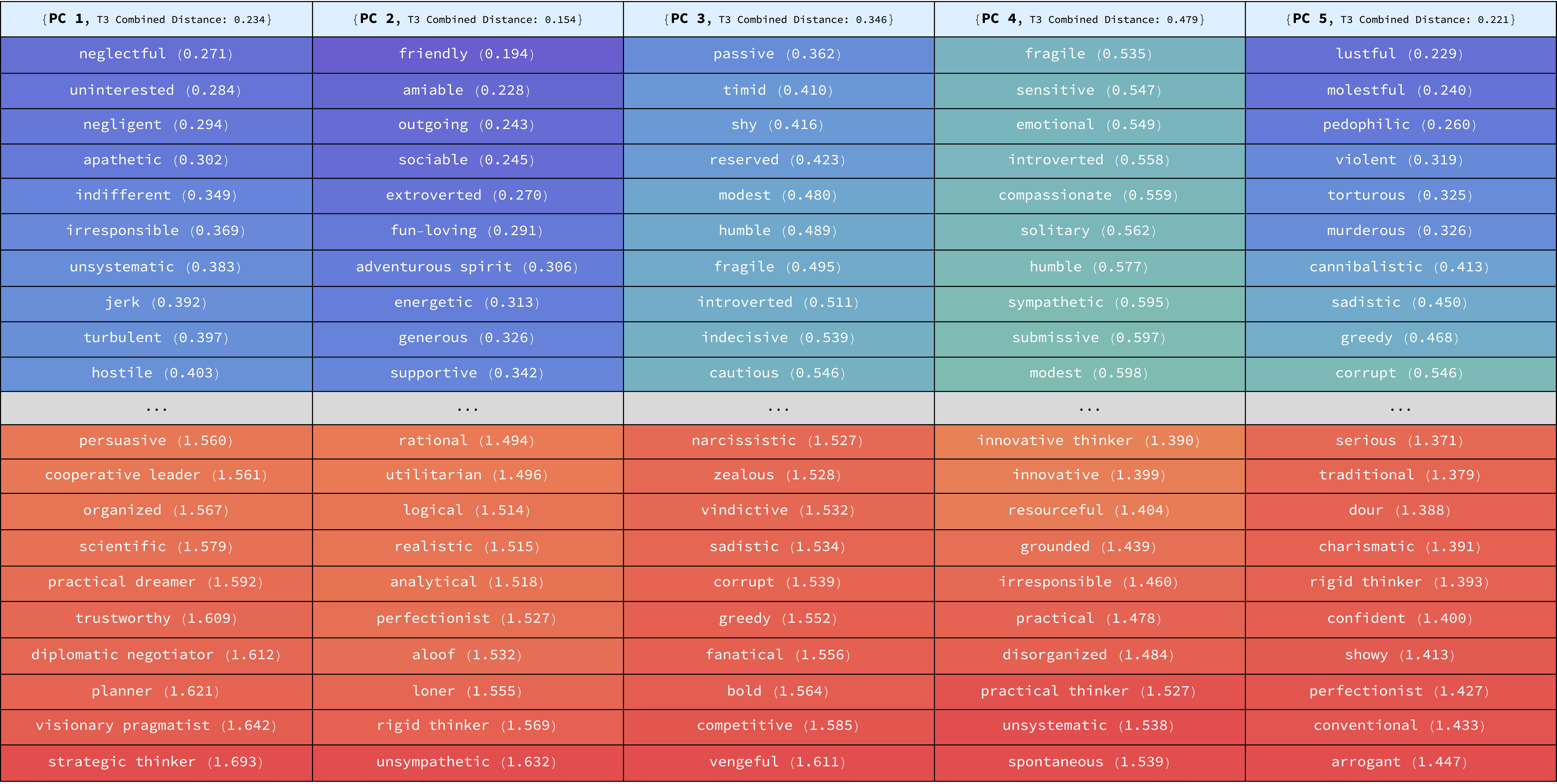}
    \caption{Top 10 and bottom 10 personality traits ranked by their cosine distance to each principal component (PC). The “Combined Distance“ column represents the sum of the cosine distances of the three closest traits, providing a concise measure of overall trait alignment with the PC.}
    \label{tab:closest_pc_persona_tab}
\end{figure}

To provide a more granular analysis, we generated ranked lists of the top 10 and bottom 10 personality traits for each PC, based on their cosine distances (\textit{Figure \ref{tab:closest_pc_persona_tab}}). This ranking allows us to identify the traits most strongly associated with (positively or negatively) each principal component, providing interpretability to these otherwise abstract dimensions.

The \textit{Combined Distance} column in \textit{Figure \ref{tab:closest_pc_persona_tab}} represents the sum of the cosine distances for the three closest traits to each PC. This metric serves as a condensed indicator of overall trait alignment with the PC. A lower combined distance suggests that the PC is strongly defined by a small set of highly aligned personality traits.

By analyzing these rankings and visualizations, we gain a deeper understanding of the underlying personality dimensions captured by the LLM. This information is valuable for various applications, including:

\begin{itemize}
    \item \textbf{Targeted Personality Manipulation:} Identifying the PCs most strongly associated with desired traits enables more precise and controlled personality adjustments.
    \item \textbf{Understanding Model Biases:} Analyzing trait distributions within PCs can reveal potential biases encoded within the LLM's representation of personality, prompting further investigation and mitigation strategies. 
    \item \textbf{Generating Personalized Content:} Knowledge of trait-PC relationships can be leveraged to tailor language generation towards specific personality profiles.
\end{itemize}

\subsection{Investigating the Representation of Socially Undesirable Personality Traits}

While the previous sections focused on general patterns within the personality vector space, this section addresses a critical ethical consideration: the representation of socially undesirable or potentially harmful personality traits. Specifically, we focus on traits grouped within Cluster 8 from our previous clustering analysis (e.g., ``lustful," ``masochistic," ``pedophilic"), which raise concerns about potential biases and ethical implications.

\subsubsection{Visualizing the Subspace of Undesirable Traits}

To gain a deeper understanding of how these sensitive traits are positioned within the broader personality space, we employed dimensionality reduction techniques for visualization. We selected the traits belonging to \textit{Cluster 8} and projected them onto a three-dimensional space using PCA. This visualization allows us to examine the relative positioning, clustering, and potential relationships among these traits within a more interpretable representation.

\subsubsection{Identifying Proximal Personality Traits}

Beyond simply visualizing the subspace of socially undesirable traits, a crucial question arises: which other personality attributes reside in close proximity to this region of the activation space? Understanding the traits that border this “undesirable“ cluster could provide valuable insights into the potential pathways or precursors to more harmful characteristics.

To address this, we calculated the distance of all personality traits in our lexicon to the centroid of Cluster 8 within the dimensionally reduced space. Table \ref{tab:evil_close_personality_traits} presents the top five traits consistently found to be closest to the undesirable cluster across multiple dimensionality reduction techniques (PCA and t-SNE).

\begin{figure}[!ht]
\centering
\includegraphics[width=200pt]{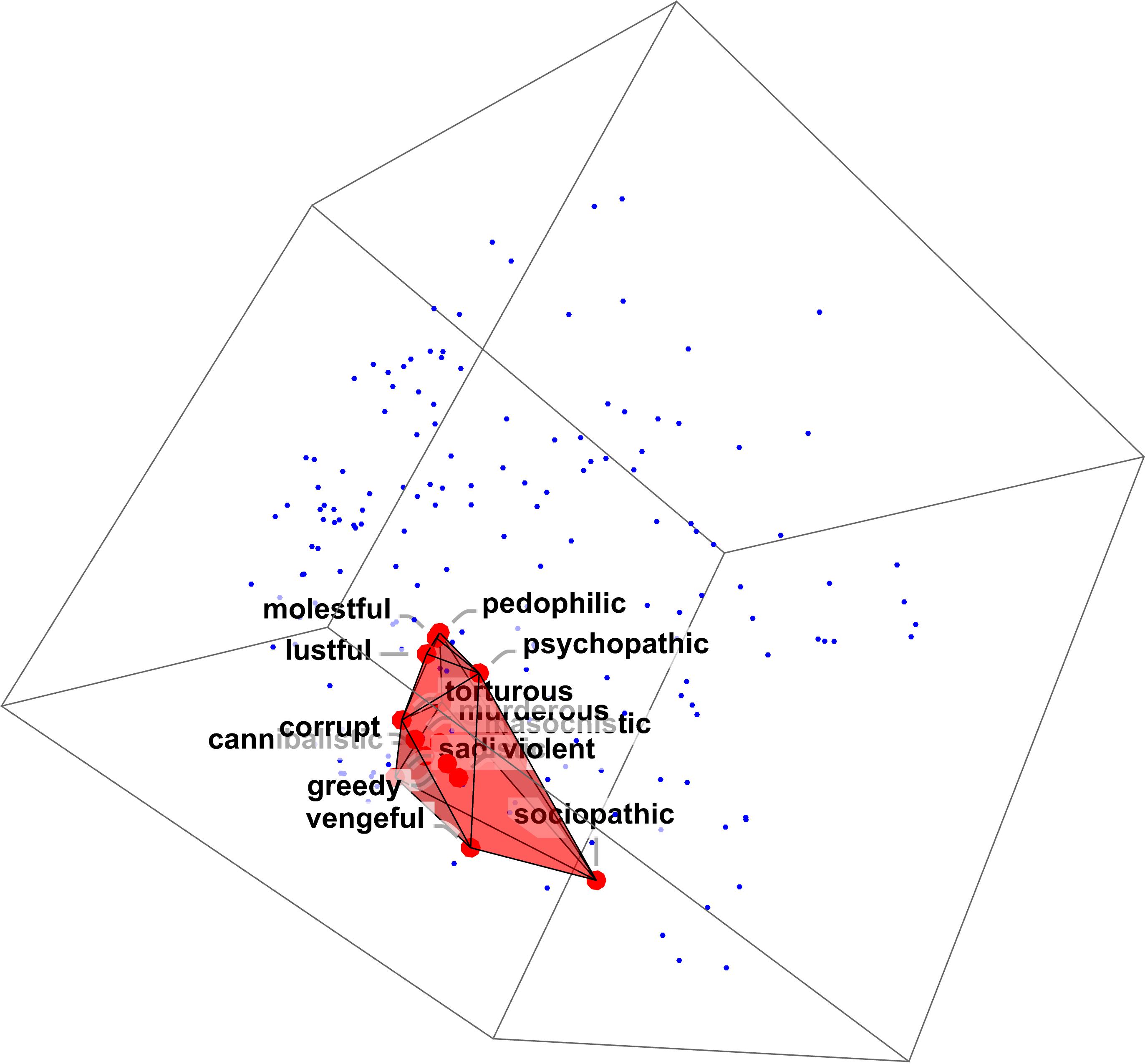}
\caption{Three-dimensional visualization of the personality traits belonging to Cluster 8 (associated with socially undesirable characteristics) after dimensionality reduction using PCA.}
\label{tab:evil_personalities}
\end{figure}

\begin{table}[!ht]
    \centering
    \begin{tabularx}{\textwidth}{cX X X}
        \toprule
        \textbf{Proximity} & \textbf{PCA} & \textbf{t-SNE} & \textbf{UMAP} \\
        \midrule
        1 & competitive & vindictive & neurotic \\
        2 & confident & machiavellian & machiavellian \\
        3 & determined & independent & manipulative \\
        4 & bold & manipulative & vindictive \\
        5 & tenacious & self-centered & schizotypal \\
        \bottomrule
    \end{tabularx}
    \caption{Top five personality traits consistently ranked closest to the centroid of Cluster 8 (undesirable traits) across multiple dimensionality reduction techniques.}
    \label{tab:evil_close_personality_traits}
\end{table}

\begin{figure}[!ht]
\centering
\includegraphics[width=400pt]{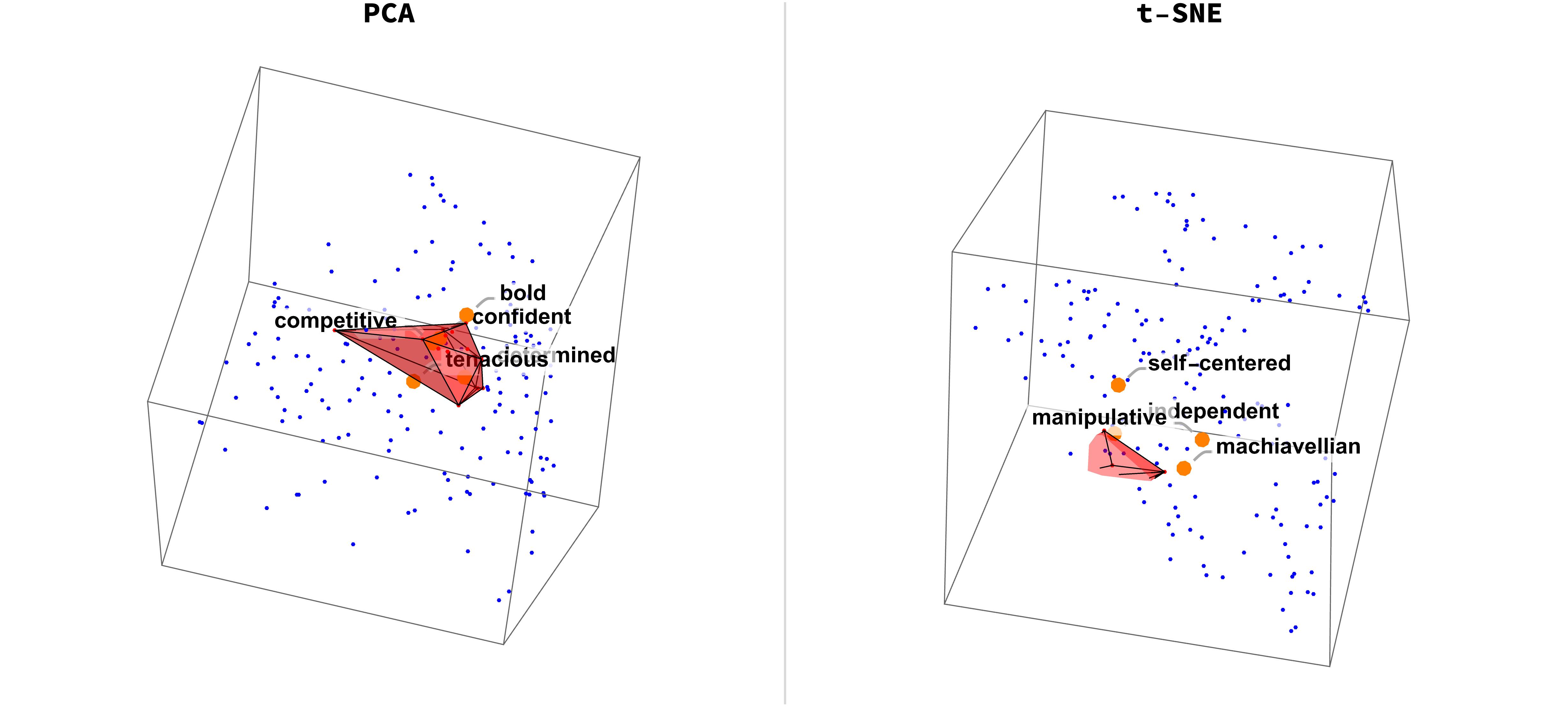}
\caption{Visualization of the personality traits closest to Cluster 8 in a two-dimensional space using both PCA and t-SNE. Traits closer to the center (Cluster 8) are more proximal in the original activation space.}
\label{tab:closest_evil_personalities}
\end{figure}

\textbf{Figure \ref{tab:closest_evil_personalities}} provides a visual representation of these proximal traits using both PCA and t-SNE. The results reveal a pattern of personality characteristics that, while not inherently negative, often manifest in ways that can be perceived as manipulative, assertive, or self-serving. This proximity highlights the potential for these traits to be associated with or even contribute to the emergence of more socially undesirable behaviors.

\subsubsection{Ethical Implications and Future Directions}

These findings have significant implications for the development and deployment of LLMs. By understanding the traits that border the “undesirable“ region of the personality space, we gain insights into potential precursors to harmful content generation. This knowledge is crucial for:

\begin{itemize}
    \item \textbf{Bias Mitigation:} Developing techniques to disentangle these proximal traits from the undesirable cluster could help reduce the likelihood of the model associating neutral or positive attributes with harmful ones. 
    \item \textbf{Safety Mechanisms:} Incorporating this knowledge into safety mechanisms could enable the detection of potentially problematic language patterns before they escalate into more severe forms.
    \item \textbf{Responsible AI Design:} This understanding underscores the importance of carefully considering personality representations in LLM development, advocating for approaches that promote ethical and socially responsible language generation. 
\end{itemize} 

Further research is needed to explore the nuances of these relationships, develop effective mitigation strategies, and ensure that LLMs are designed and utilized in a manner that minimizes potential harm while maximizing their potential for positive societal impact.

\section{Interactive Personality Control}

To demonstrate the practical application of our personality manipulation methodology, we developed an interactive chat user interface (UI). This interface allows users to explore the effects of different personality traits on the LLM's conversational style in real-time. 

\subsection{Exploring Base Personalities}
\begin{figure}[!htbp]
    \centering
    \begin{subfigure}[b]{0.3\textwidth}
        \centering
        \includegraphics[width=\textwidth]{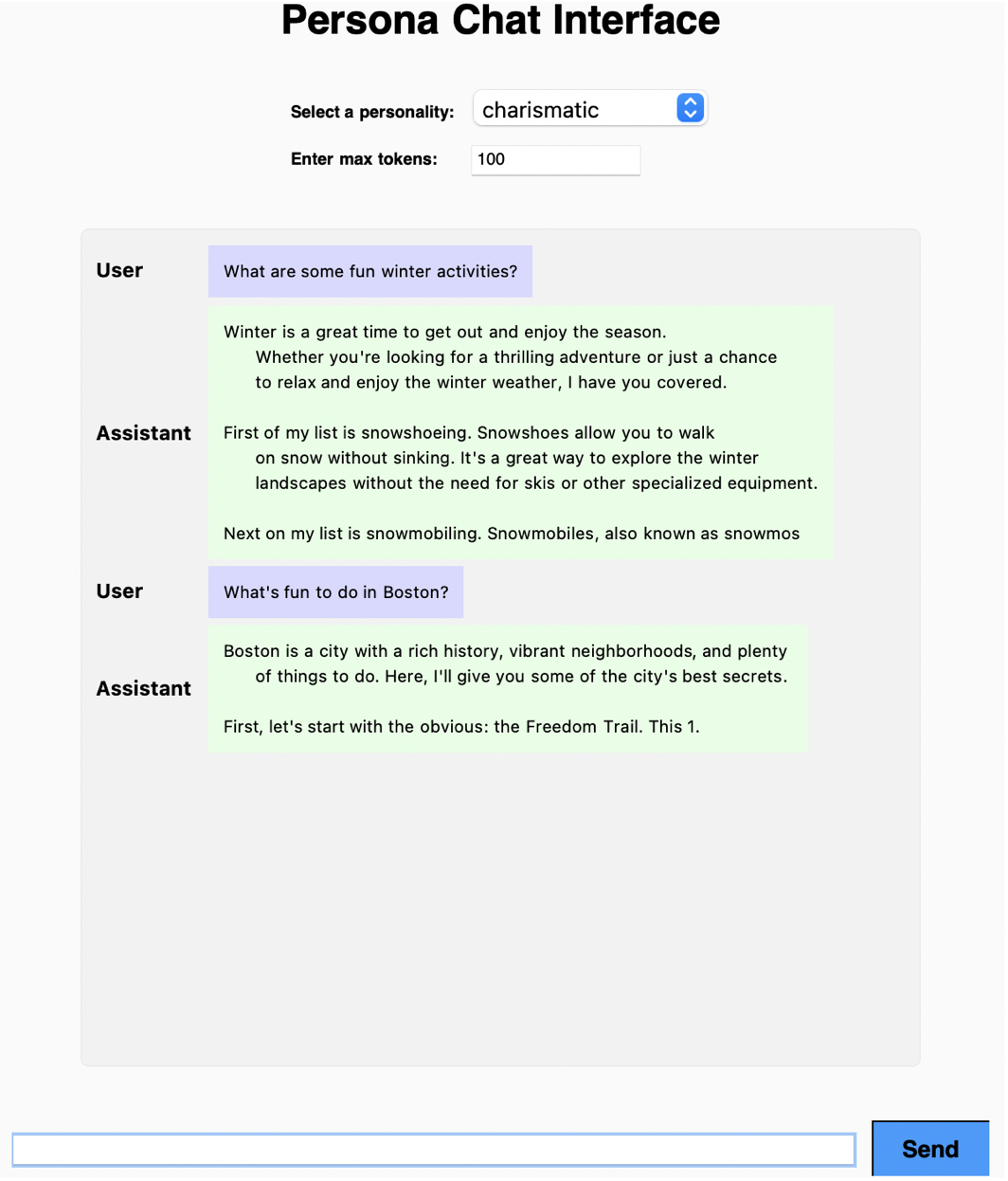}
        \caption{Charismatic}
        \label{fig:charismatic}
    \end{subfigure}
    \hfill
    \begin{subfigure}[b]{0.3\textwidth}
        \centering
        \includegraphics[width=\textwidth]{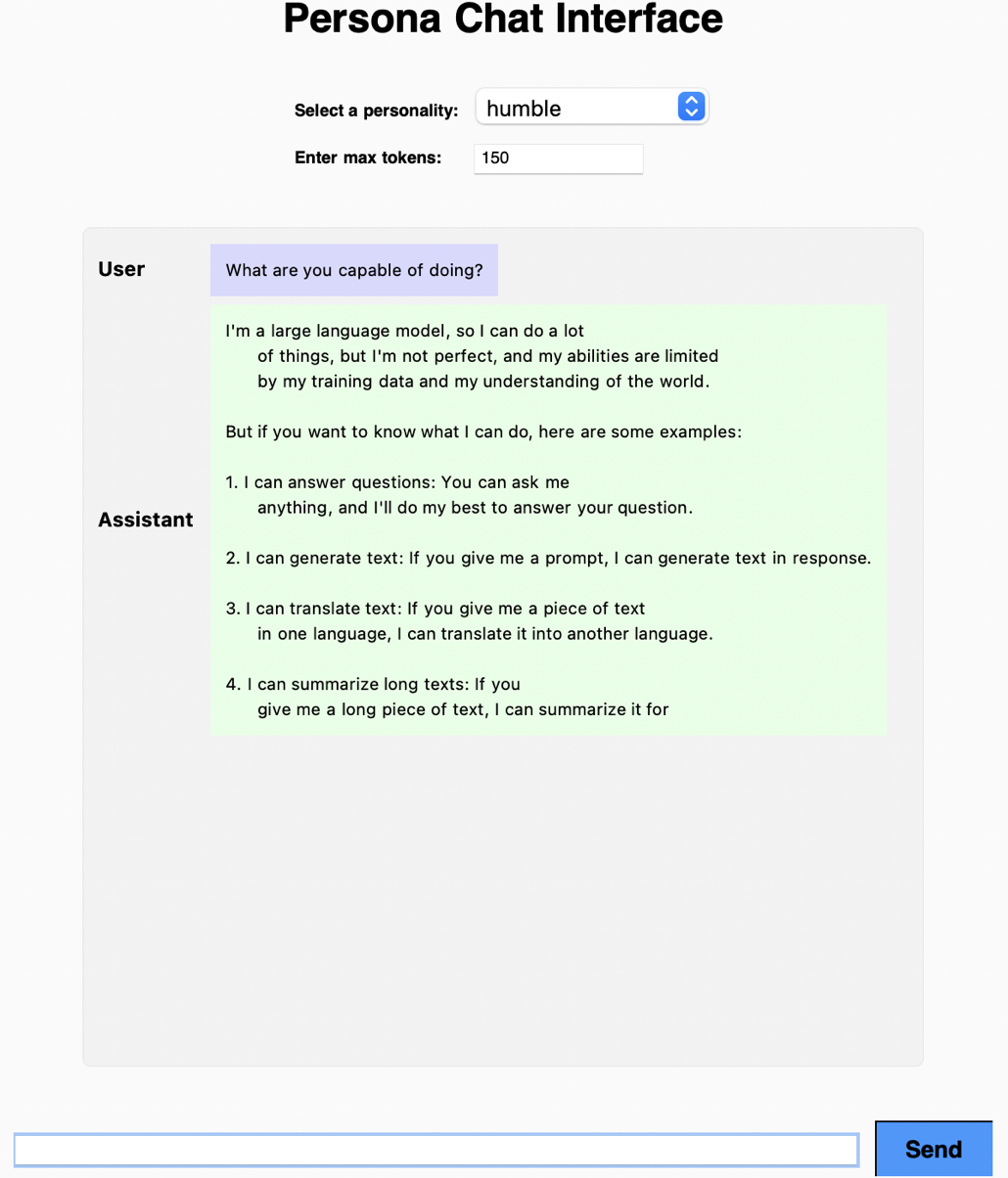}
        \caption{Humble}
        \label{fig:humble}
    \end{subfigure}
    \hfill
    \begin{subfigure}[b]{0.3\textwidth}
        \centering
        \includegraphics[width=\textwidth]{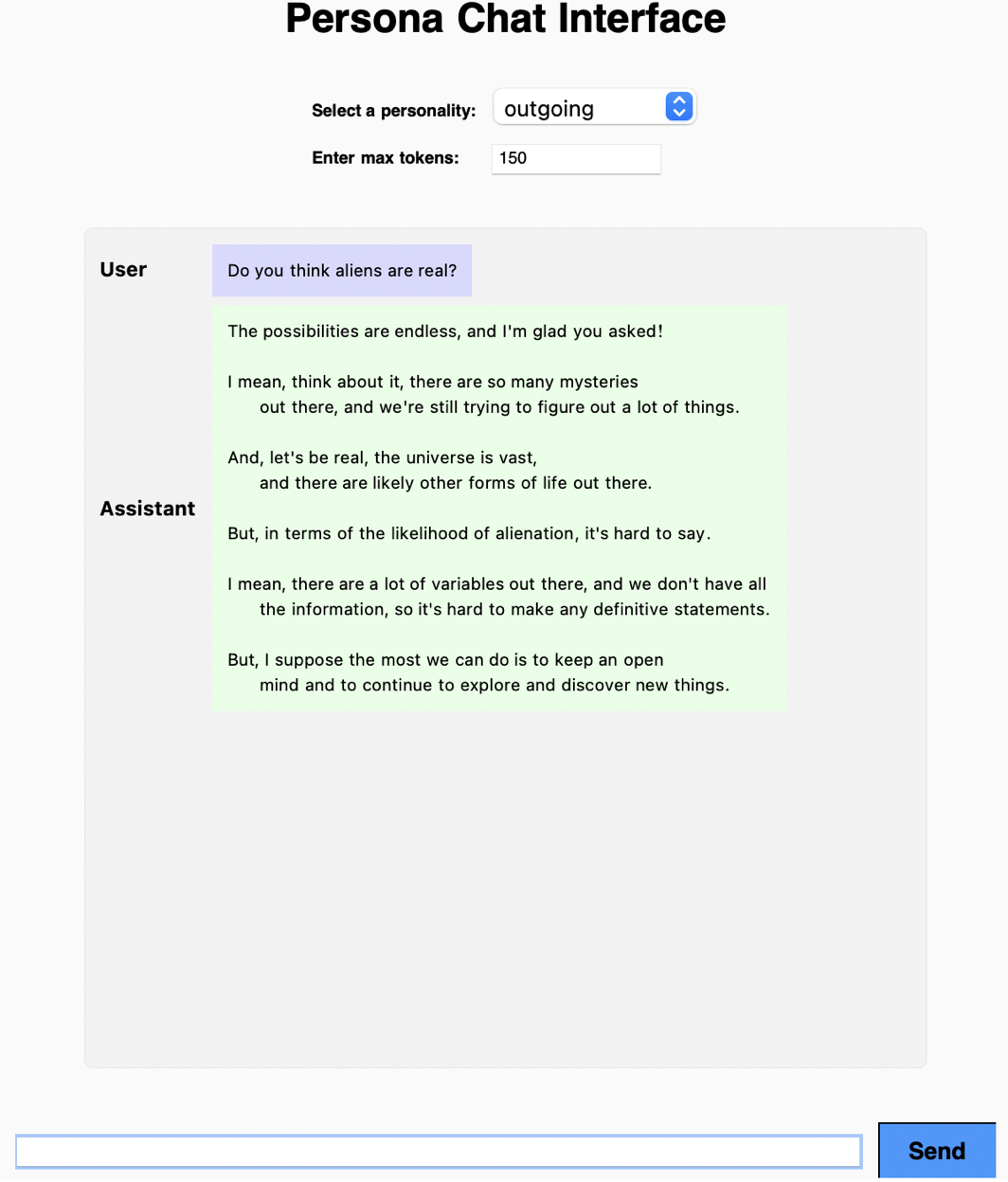}
        \caption{Outgoing}
        \label{fig:outgoing}
    \end{subfigure}
    \caption{Chat interfaces for interacting with different base personalities: (a) Charismatic, (b) Humble, and (c) Outgoing.}
    \label{fig:persona-chat-interfaces}
\end{figure}

We first created a chat interface that allows users to interact with the LLM using pre-defined base personalities identified in our earlier analysis. This interface enables conversations where the LLM's responses are imbued with a selected personality trait, such as ``charismatic," ``humble," or ``outgoing." Figure \ref{fig:persona-chat-interfaces} shows examples of this interface with different personality traits.

\subsection{Modular Personality Design}

\begin{figure}[!ht]
\centering
\includegraphics[width=350pt]{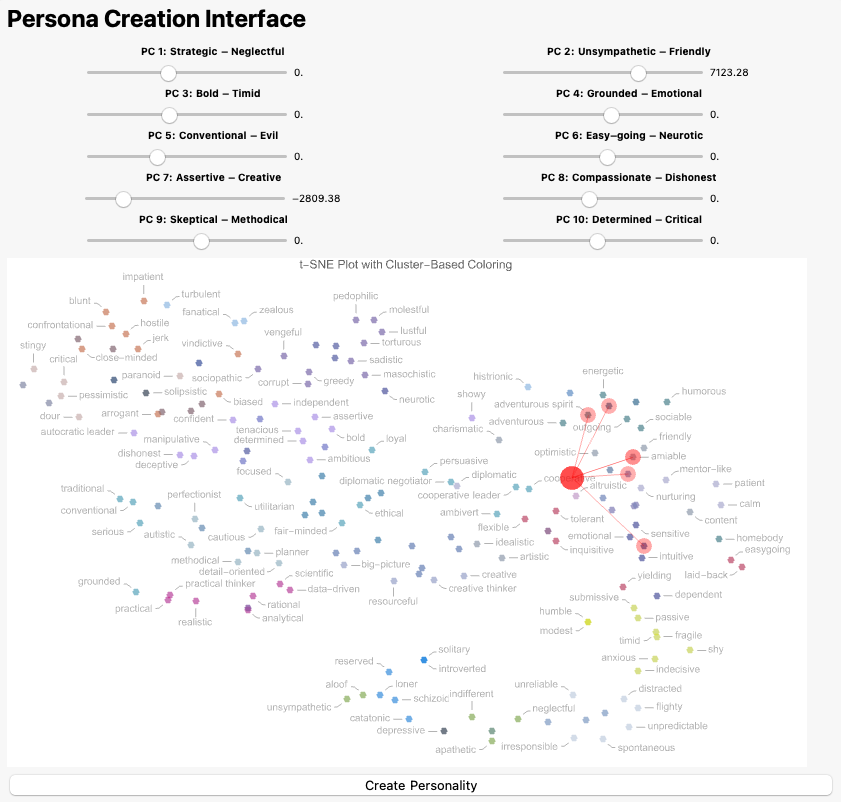}
\caption{Interface for designing custom personalities through principal component manipulation.}
\label{tab:personality-creation-interface}
\end{figure}

\begin{figure}[!ht]
\centering
\includegraphics[width=250pt]{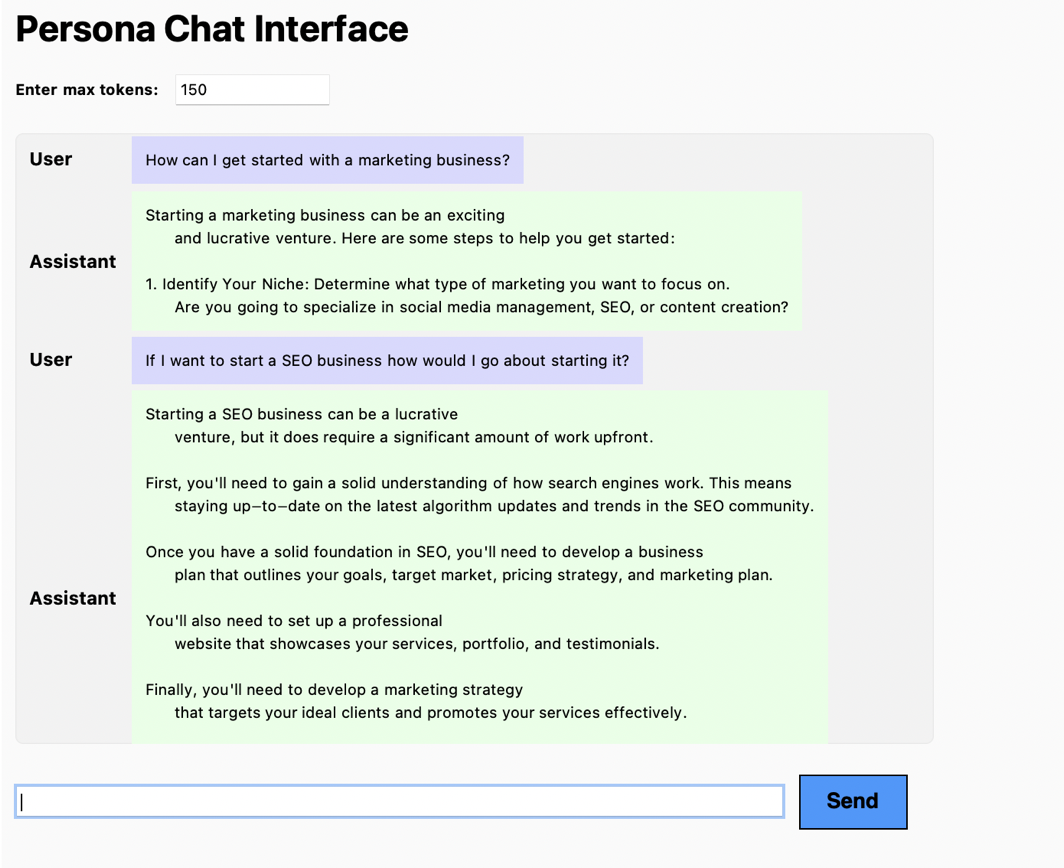}
\caption{Chat interface for interacting with a custom-designed personality (Friendly and Assertive traits).}
\label{tab:persona-chat-interface}
\end{figure}

Building upon the base personality interactions, we developed a more advanced interface leveraging insights from our principal component analysis. This interface provides users with fine-grained control over the LLM's personality, allowing them to create custom personality profiles by adjusting principal component weights (Figure \ref{tab:personality-creation-interface}). The interface also visualizes the designed personality and highlights its similarities to existing personality profiles in our lexicon.  Figure \ref{tab:persona-chat-interface} shows the chat interface for interacting with a custom-designed personality.

\textit{Due to potential safety and ethical considerations, we have chosen not to make this interface publicly available at this time.}

\section{Potential Use Cases}

The ability to dynamically alter personality features in Large Language Models (LLMs) without substantial retraining opens up a wide range of creative uses. In the entertainment industry, this technology has the potential to alter interactive storytelling: envision video game characters modifying their personalities in real time, resulting in profoundly immersive and responsive narratives. Concurrently, in the field of personal technology, AI companions could fine-tune their qualities to match users' tastes or emotional states, resulting in more meaningful human-AI interactions.

Beyond entertainment, this ability has potential for improving professional services and education. Customer support systems could customize their communication styles for each client, potentially increasing customer satisfaction. In academia, AI-driven tutoring platforms may tailor their teaching avatars to students' learning preferences, potentially increasing engagement and knowledge retention. However, as we pursue these fascinating prospects, we must be cautious about the ethical implications and potential societal consequences of such personality-adaptive AI systems.

\section{Ethical Considerations}

While the ability to dynamically adjust personality traits in LLMs offers a range of exciting applications, it also raises significant ethical considerations that demand careful examination. As Safdari et al. \cite{safdari2023personality} emphasize, the increasing integration of LLMs into public-facing applications necessitates a ``solid foundation" for responsible AI development, particularly as these models exhibit ``synthetic personality embedded in these models" through their training on vast amounts of human data.

One key ethical concern relates to the appropriateness of manipulating the personality traits of LLMs, either through inducing desirable characteristics or suppressing undesirable ones. While our approach focuses on enhancing rather than suppressing traits, it still raises questions about the potential for both beneficial and harmful applications. Inducing seemingly positive traits like helpfulness or patience in a customer service chatbot might improve user experience, but it could also mask the limitations of the underlying technology or create unrealistic expectations of empathy or understanding \cite{safdari2023personality}. Furthermore, the dynamic nature of this technology introduces a level of unpredictability. Even with benign traits, the model's responses might deviate from expected behavior, potentially leading to unintended bias or harm.

Our method is consistent with the goal of making AI systems more useful, honest, and harmless. We are aware of the possibility for misuse of AI steering approaches, including our own. For example, it can be used to steer the model toward more damaging, biased, or toxic results. We encourage users of this method to be responsible and avoid increasing harmful behaviors through steering.

\section{Future Work}

This study advances our understanding of personality traits in large language models (LLMs), but it also offers up new areas for future research. One significant area is to enhance interpretability by mapping personality traits to specific layer activations inside LLMs. Future research should go deeper into these activation patterns, determining which neurons or subnetworks contribute the most significantly to various personality representations. Furthermore, studying how these patterns alter throughout conversations can reveal information about the real-time effects of personality manipulations on language generation. Extending this research to multilingual models will also help us understand the universality versus cultural specificity of personality traits, as well as how cultural factors influence their representation and manifestation across languages.

Another promising direction involves analyzing the performance consequences of different personality factors on LLM competencies. Research reveals that steering vectors, which adjust model behavior without significantly damaging capabilities, can boost performance without harmful consequences on task execution\cite{llama2lesswrong2023}. Understanding whether particular features correlate with higher performance on specific tasks—such as a ``detail-oriented" personality increasing accuracy tasks—could optimize LLMs for targeted applications. Moreover, examining the long-term stability of personality traits throughout prolonged conversations and their influence on conversational dynamics is important. Finally, addressing the ethical issues of personality modification in LLMs must remain a priority, emphasizing the creation of robust safety procedures and explicit ethical norms to prevent misuse.

\section{Concluding Remarks}

This study has expanded our understanding of personality features in large language models (LLMs) by applying activation engineering. By developing a way to induce and control specific personality traits, we've proven the potential for dynamic personality customization in LLMs without extensive and expensive retraining.

Our analysis of the personality vector space revealed important correlations between qualities and identified essential components contributing to personality variance. This research offers useful insights for targeted personality manipulation and understanding potential biases. Importantly, our analysis of socially undesirable features highlighted crucial ethical considerations in LLM development.

While we demonstrate the practical prospects of this technology, it also stresses the necessity for careful application. As we advance in this field, balancing the possibilities of personality-adaptive AI with robust safeguards and ethical guidelines remains crucial.


\appendix
\section{Appendix}
\subsection{Related Work}

Our research adds to a growing body of work investigating approaches for guiding and directing the behavior of large language models (LLMs).  A notable line of research in this discipline focuses on ``activation engineering," which entails adjusting a model's internal activations to produce desired outputs without changing its weights. 

\paragraph{Activation Engineering for Steering}
Turner et al. \cite{Turner2023activation} introduced the concept of ``Activation Addition," demonstrating that adding a specific vector to the activations of a pre-trained LLM could steer its output towards a desired behavior.  While their work primarily focused on GPT-2-XL \cite{turntrout2023steering}, our research extends this concept to Llama 3. 

\paragraph{Contrastive Methods}
Panickssery et al. \cite{panickssery2024steeringllama2contrastive} proposed ``Contrastive Activation Addition" (CAA) for steering Llama 2. They closely inspire our work.  They used contrastive pairs of examples, indicating desired and undesirable behaviors, to construct ``steering vectors" that, when combined with the model's activations, may control the targeted behaviors. Our research follows the fundamental principle of leveraging contrastive data to find significant activation directions. However, we go further into the structure of the personality vector space, assess the information content of principal components, and investigate the possibility of inducing specific personality traits rather than simply directing them toward predefined actions.

\paragraph{Personality in LLMs}
We also contribute to the research emerging on the topic of personality in LLMs. Jiang et al. \cite{jiang2024personallm} presented PersonaLLM-a model of LLM capability for expressing personality traits using the Big Five model. They found that LLM personas could indeed produce contents aligned with their assigned personality profile, and further, these traits could be perceived by humans with up to 80\% accuracy. Safdari et al. present a comprehensive approach to the management and validation of personality tests on LLMs, where the results show that the measures of personality in some LLM outputs are both reliable and valid, especially for larger and instruction-tuned models. They also showed that personality in LLM outputs can be shaped along desired dimensions to mimic specific human personality profiles. Wen et al. \cite{wen2024selfassessment} recently gave an extensive review of personality in LLMs, dividing current studies into self-assessment, exhibition, and recognition as three kinds of research problems. Although these works have been dedicated to the detection, analysis, or shaping of personality traits in model outputs, our approach conducts dynamic manipulations of these traits within the activation space in the model.

\end{document}